\documentclass[10pt,twocolumn,letterpaper]{article}
\usepackage{graphicx}

\usepackage[pagenumbers]{wacv} 

\definecolor{wacvblue}{rgb}{0.21,0.49,0.74}
\usepackage[pagebackref,breaklinks,colorlinks,allcolors=wacvblue]{hyperref}

\def\wacvPaperID{104} 
\def\confName{WACV}
\def\confYear{2027}

\title{Pano2World: End-to-End 3D Generation via Unified Multi-View Sequences}

\author{Zhenjia Li\\
Ke Holdings Inc.\\
Beijing, China\\
\and
Jinrang Jia\\
Ke Holdings Inc.\\
Beijing, China\\
\and
Yifeng Shi\\
Ke Holdings Inc.\\
Beijing, China\\
}

\begin{document}
\maketitle

\begin{figure*}[!t]
  \centering
  \includegraphics[width=\textwidth]{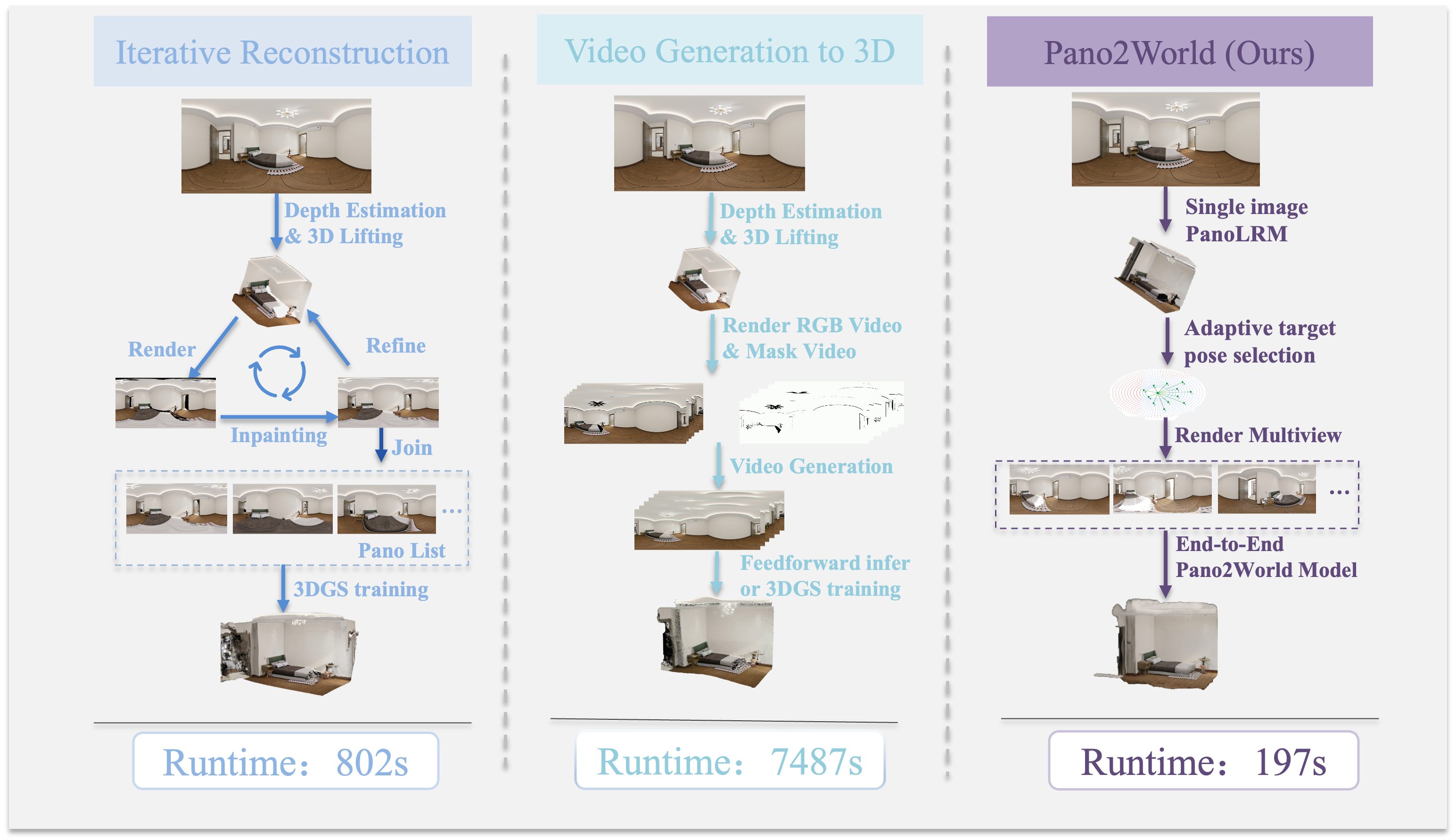}
    \caption{
    Pipeline comparison for single-panorama 3D scene generation.
    Pano2World avoids the repeated scene updates required by iterative reconstruction methods and does not rely on continuous trajectory video generation used in video-to-3D pipelines.
    Instead, it generates only the necessary multi-view observations at adaptively sampled target poses and directly predicts the final 3DGS~\cite{kerbl2023gaussian}.
    The runtime shown at the bottom highlights the efficiency advantage of Pano2World in end-to-end generation.
    }
  \label{fig:qwen-mvp-case}
\end{figure*}

\begin{abstract}
A single panorama captures the full visual sphere from one camera center, yet confines users to looking around in place without enabling true scene exploration.
Converting a single panorama into a persistent, renderable 3D representation for free-viewpoint navigation has attracted growing interest; existing methods either adopt iterative per-view completion that propagates inpainting results to update the underlying geometry, leading to progressive error accumulation and cumbersome multi-step pipelines, or leverage the temporal consistency priors of video generation models, yet the continuous-trajectory constraint intrinsic to such models limits their flexibility in covering scenes from multiple directions simultaneously.
We present \textbf{Pano2World}, which takes a single indoor panorama as input and directly outputs a persistent, explorable 3D Gaussian scene.
Given the source panorama, Pano2World first reconstructs a coarse 3D Gaussian proxy and renders it at adaptively sampled nearby poses to obtain geometrically aligned guidance panoramas; a panoramic diffusion model then jointly denoises all target views via \textit{View-Aware Attention Routing}, where each target view simultaneously receives geometric constraints from its corresponding guidance panorama and global semantic guidance from the source panorama, naturally enforcing cross-view consistency.
To avoid the information loss incurred by decoding the multi-view hidden features formed during joint denoising back to the pixel domain via VAE, we introduce Latent Feature Adapter, a geometry-aware bridge module that directly distills these hidden features into a scene latent, subsequently decoded into the final 3D Gaussian scene.
Experiments demonstrate that Pano2World significantly outperforms existing methods on the multi-position panoramic novel-view synthesis benchmark.
\end{abstract}

\section{Introduction}

Compared with perspective images with limited fields of view, a single indoor equirectangular panorama captures the complete visual sphere from one camera center and provides rich global layout and appearance information. It has therefore been widely used in applications such as real-estate visualization. However, a single panorama remains a single-center observation: it contains no translational parallax~\cite{jia2023monouni} and thus cannot directly support the local camera motion required by emerging applications such as VR walkthroughs and immersive scene browsing. To move beyond this fixed-viewpoint limitation, it is necessary to construct a 3D scene representation like 3D Gaussian Splatting(3DGS)~\cite{kerbl2023gaussian,xing2026adaptsplat,jia2026yogo} from a single panorama that supports rendering from nearby viewpoints.

Although directly generating a 3D scene conditioned on an input image is an appealing direction, existing generative models remain constrained by the lack of large-scale scene-level 3D data. As a result, most existing approaches rely on 2D image or video priors to indirectly construct scene representations. One line of methods estimates an initial geometry from the source panorama and progressively completes the scene through an iterative process of rendering, image inpainting, and geometry update~\cite{pu2024pano2room}. However, such per-view processing can propagate early completion errors into subsequent geometry updates, and each target view requires repeated rendering and inpainting, limiting overall efficiency. Another line of methods leverages the temporal consistency of video generation models to synthesize multi-view observations, which are then converted into 3D representations through reconstruction modules~\cite{hyworld2026hyworld2,yang2025matrix3d}. While these methods benefit from strong generative priors, their generation process is typically organized around a single continuous trajectory, making it difficult to naturally cover target poses that expand from the source viewpoint in multiple directions. They also introduce many intermediate frames that are unnecessary for the final local exploration task. Therefore, generating geometrically consistent multi-position observations while avoiding both per-view iteration and redundant trajectory generation remains a key challenge for single-panorama scene exploration.

To address these limitations, we reformulate single-panorama scene expansion as a joint generation problem over multiple target poses. Instead of sequentially completing each novel view, we directly synthesize panoramic observations at a set of nearby target poses in parallel. This design exploits the strong priors of image generation models while avoiding the error accumulation caused by per-view iteration. By explicitly modeling target camera poses and introducing cross-view consistency constraints, the model can preserve the flexibility of target-pose layouts while producing multi-view observations that are better suited for subsequent 3D scene reconstruction. In addition, we introduce a geometry-aware bridge module that directly converts DiT features into features required by the 3D representation, improving 3D consistency while further reducing generation overhead.

Based on this idea, we propose Pano2World, a framework that directly generates a locally explorable 3DGS scene from a single indoor panorama. Pano2World first predicts a coarse 3D Gaussian proxy scene from the source panorama and renders geometrically aligned but incomplete guidance panoramas at a set of nearby target poses. These guidance panoramas provide pose-specific local geometric anchors for each target view, while the source panorama supplies complete scene appearance and semantic context. We then employ PanoDiT, a multi-view panoramic generation model, to jointly complete all target views. To reduce the computational cost of unconstrained full-token interactions while strengthening the geometric binding between each target view and its corresponding guidance panorama, we design View-Aware Attention Routing. Furthermore, we introduce a cross-view consistency constraint to improve geometric coordination among the generated panoramas, providing more reliable multi-view observations for subsequent unified 3DGS prediction.
Unlike prior pipelines that decode image-generation latents into RGB images before reconstructing a 3D scene, our framework directly exploits the hidden features formed during multi-view joint denoising. Since PanoDiT jointly denoises multiple target views, its hidden features encode rich cross-view interaction information. We therefore extract the DiT hidden features before PanoDiT's final output projection, fuse them with the corresponding camera-ray geometry, and transform them through the Latent Feature Adapter, a geometry-aware bridge module, into the scene latent required by the Gaussian Decoder~\cite{jia2026panoworld}. The decoder then produces the final 3DGS. By bypassing RGB decoding and re-encoding, this design reduces additional computation and more faithfully preserves the cross-view structural information formed during joint generation, thereby improving the consistency and renderable quality of the final 3DGS scene.

In summary, this paper makes three main contributions:
\begin{itemize}
\item We propose Pano2World, which formulates single-panorama 3D generation as a multi-position joint generation problem. By directly leveraging latent-domain PanoDiT hidden features to drive 3D prediction, Pano2World converts a single indoor panorama into a freely explorable 3DGS scene within minutes.
\item We propose PanoDiT, a jointly denoising multi-view panoramic completion framework. By incorporating View-Aware Attention Routing and a cross-view consistency loss, PanoDiT enables efficient generation of multi-view images with strong 3D consistency.
\item Comprehensive experimental results and ablation studies demonstrate that Pano2World achieves superior novel-view rendering quality, cross-view consistency, and generation efficiency compared with existing methods, while validating the effectiveness of the proposed modules.
\end{itemize}

\section{Related Work}

\subsection{Iterative Scene Expansion from a Single Image or Panorama}

Early methods for constructing explorable 3D scenes from a single image or panorama commonly adopt an iterative scene expansion pipeline. These methods first estimate an initial geometry, render the current scene representation from a novel viewpoint, inpaint holes or disoccluded regions, and fuse the completed result back into the scene representation. In the panoramic domain, Pano2Room applies this warp-and-inpaint strategy to novel view synthesis from a single indoor panorama~\cite{pu2024pano2room}; related single-image scene generation methods also rely on geometry scaffolds or intermediate scene representations to support progressive expansion~\cite{yu2024wonderworld,yao2026anchoreddream}. However, such sequential pipelines limit cross-view information exchange, and early inpainting errors may propagate through subsequent geometry updates. In addition, rendering, inpainting, and geometry update must be repeatedly executed as the number of target views increases. In contrast, Pano2World formulates nearby scene expansion as a joint generation problem over multiple target poses, avoiding per-view iteration and improving inference efficiency.

\subsection{Multi-View and Panoramic Generative Priors}

Diffusion models have substantially improved the quality of image, multi-view image, and panoramic image generation. Multi-view diffusion methods~\cite{tang2024mvdiffusionpp,li2024era3d,gao2024cat3d} synthesize consistent multi-view observations from a single image or sparse views, while panoramic generation methods such as~\cite{zhang2024panfusion,liu2024panofree,ye2024diffpano,kalischek2025cubediff} study 360-degree image synthesis, spherical geometry modeling, and cross-view consistency for wide-field-of-view representations. These methods provide strong 2D generative priors for appearance completion and semantic synthesis, but they are typically designed for image generation, object-level reconstruction, or text-conditioned panorama synthesis rather than constructing a persistent, locally explorable 3D scene from a single indoor panorama. Pano2World instead synthesizes target-pose-aligned panoramic observations and connects the latent representations formed during generation to a 3DGS decoder through a geometry-aware bridge.

\subsection{World-Consistent Video Generation}

Video diffusion models have recently been widely explored for controllable view synthesis and world-consistent generation. Some methods~\cite{markyu2025trajectorycrafter,yu2024viewcrafter} condition video generation on camera trajectories, expanding a single image into a view sequence along a specified path. Subsequent methods introduce 3D geometric constraints, long-term spatial memory, or local spatial retrieval mechanisms to mitigate geometric drift, spatial forgetting, and appearance inconsistency in long-horizon generation~\cite{ren2025gen3c,xu2025geometrycrafter,wu2025videoworldmemory,wang2026anchorweave}. However, video generation is still organized as a temporally ordered sequence of continuous frames. For local exploration from a single panorama, the desired output is a set of target poses expanding from the source viewpoint in multiple directions, rather than densely sampled frames along a single trajectory. Directly adopting video generation therefore introduces redundant intermediate frames and requires an additional video-to-3D reconstruction stage~\cite{jia2026yogo,liu2025attentiongs,wang2026artifactworld}. In contrast, Pano2World jointly generates panoramic observations directly at the target pose set and transforms the latent features formed during joint generation into a unified 3DGS representation.

\section{Method}

\begin{figure*}[t]
  \centering
  \includegraphics[width=\textwidth]{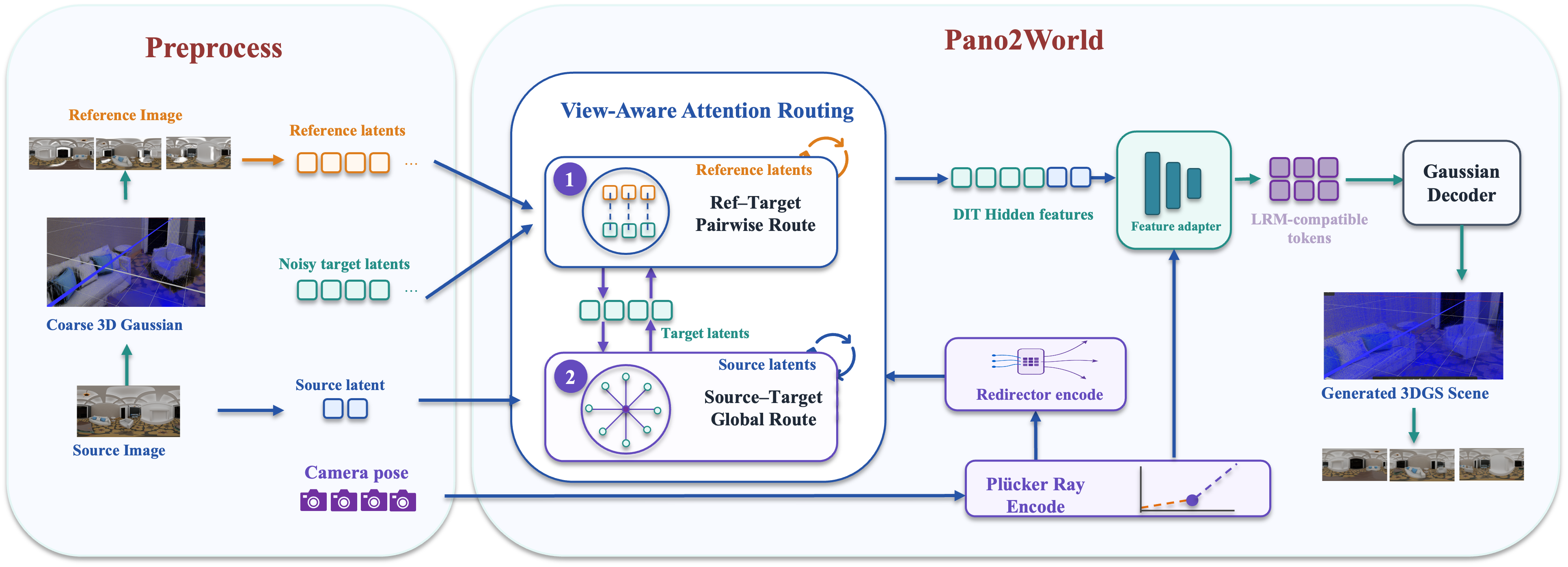  }
    \caption{
    Architecture of Pano2World.
    The source panorama is first lifted to a coarse 3D Gaussian proxy, which is rendered at target poses to provide geometrically aligned reference panoramas.
    PanoDiT jointly completes multiple target views with View-Aware Attention Routing and ReDirector camera conditioning.
    The Latent Feature Adapter maps the denoising hidden features and camera-ray geometry into LRM-compatible tokens, which are then decoded into the final renderable 3D Gaussian scene.The detailed adaptive target-pose sampling strategy is provided in the supplementary material.
    }
  \label{fig:architecture}
\end{figure*}

Given a single indoor equirectangular panorama $I_s$ captured at camera center $\mathbf{c}_s$ with pose $P_s \in \mathrm{SE}(3)$, the objective is to predict a persistent 3D Gaussian scene $G$ from this single
  observation that supports free-viewpoint rendering within a local neighborhood of $\mathbf{c}_s$.

\subsection{Method Overview}

Given the source panorama $I_s$, Pano2World first feeds it into a feed-forward panoramic reconstruction model~\cite{jia2026panoworld} to predict a coarse 3D Gaussian scene $G_0$, adaptively samples a set of target poses ${P_1, \ldots,
P_N}$ according to the depth information of $G_0$, and renders $G_0$ at these poses to obtain geometrically aligned guidance panoramas ${\hat{I}_1, \ldots, \hat{I}_N}$, providing geometric anchors for subsequent
generation. As illustrated in Figure~\ref{fig:architecture}, the core network consists of three modules: PanoDiT for image-domain multi-view completion, a geometry-aware feature bridge connecting the
feature spaces of the two modules, and a Gaussian Decoder for 3D scene prediction. The detailed adaptive target-pose selection strategy based on the depth rendered from the coarse 3DGS proxy is described in the supplementary material.

\subsection{PanoDiT}

PanoDiT is built on Qwen-Edit and adapted for panorama-conditioned multi-view generation via LoRA training. The source panorama $I_s$, guidance panoramas ${\hat{I}_i}$, and noisy target latents are encoded
by a VAE and packed into a unified latent sequence, which is jointly denoised by a flow-matching DiT. However, applying unconstrained global attention over the packed sequence introduces substantial unnecessary
computation from irrelevant token interactions, and fails to guarantee precise geometric binding between each target view and its corresponding guidance panorama. To address this, we propose \textbf{View-Aware
Attention Routing}, which decomposes image token interactions into two complementary routes. The first is a \emph{reference-target pairwise route}: for the $i$-th target pose, target noisy tokens form a local
attention group exclusively with the corresponding guidance tokens, strengthening the one-to-one geometric binding between each target view and its pose-aligned guidance rendering. The second is a
\emph{source-target global route}: target tokens attend to the source panorama, inheriting scene identity, appearance style, material distribution, and room-scale semantics from the source
panorama. In practice, a block-wise schedule is applied: geometric alignment is more critical in the early denoising stage, so shallower Transformer blocks prioritize the pairwise route to establish stable local
geometric constraints; as denoising progresses and semantic fusion becomes dominant, deeper blocks switch to the global route to enhance appearance fusion between target views and the source panorama. This
design improves the stability of target-view generation while substantially reducing computational cost.

View-Aware Attention Routing constrains the scope of token interactions, but standard DiT positional encoding only describes the 2D position of tokens within images and still lacks explicit modeling of camera
poses and cross-view ray relationships. To address this, PanoDiT follows the rotary camera encoding approach of ReDirector~\cite{park2026redirector} and computes a Plücker ray descriptor for each panorama token from its
equirectangular coordinate $(u, v)$ and the corresponding camera extrinsic:
\[
g(u,v) = \bigl[\mathbf{o} \times \mathbf{d}(u,v),\; \mathbf{d}(u,v)\bigr] \in \mathbb{R}^6 .
\]
where $\mathbf{o}$ is the camera center and $\mathbf{d}(u,v)$ is the world-space ray direction. This descriptor is passed through a lightweight camera projector to produce a token-wise camera embedding that
predicts additional RoPE phase shifts in attention. The camera-conditioned phase is injected separately into the query/key path to modulate cross-view attention weights, and into the value/output path to
introduce geometry-aware feature aggregation, enabling PanoDiT to explicitly perceive camera rays and relative pose relationships during multi-view denoising.

The training objective of PanoDiT combines a flow-matching loss $\mathcal{L}{\mathrm{flow}}$ and an image-domain perceptual loss $\mathcal{L}{\mathrm{lpips}}$ for basic supervision, together with a
cross-view depth consistency loss $\mathcal{L}_{\mathrm{warp}}$ that explicitly constrains geometric consistency across target views. For a pair of neighboring target views $(i, j)$, each pixel $(u,v)$ in view 
$i$ is back-projected to a 3D world point using its ray direction and depth, and then re-projected into view $j$ with the corresponding color sampled via differentiable grid sampling, yielding $\mathrm{Warp}{j\to i}(\hat{I}0^j)$. To exclude invalid supervision from occlusions and incorrect 
projections, the loss is computed only on the valid pixel set $\Omega_{ij}$ satisfying the depth consistency condition:
\[
\mathcal{L}_{\mathrm{warp}}^{ij} = \frac{1}{|\Omega{ij}|}\sum_{(u,v)\in\Omega_{ij}} \bigl\|\hat{I}_{0}^i(u,v) - \mathrm{Warp}_{j\to i}(\hat{I}_0^j)(u,v)\bigr\|_{1}.
\]
$\mathcal{L}_{\mathrm{warp}}$ is averaged over all valid view pairs, providing explicit cross-view geometric supervision that complements the network architecture's modeling of multi-view consistency.

\subsection{Latent Feature Adapter}

The hidden features of PanoDiT encode rich multi-view interaction information through the joint denoising process; however, these features are incompatible with the token representation space
required by the Gaussian Decoder and cannot be directly consumed. Decoding back to the image domain via VAE and re-encoding would cause the cross-view structural information formed during generation to be lost
in the pixel decoding process. To address this, we introduce the \textbf{Latent Feature Adapter}, which adapts the hidden features $H_{\mathrm{pano}} \in \mathbb{R}^{B \times L \times D_p}$ extracted before
PanoDiT's final output projection into tokens that can be directly consumed by the 3DGS Decoder.

Specifically, the Latent Feature Adapter first concatenates the Plücker ray descriptor $g(u,v)$ of each token with the corresponding hidden feature in $H_{\mathrm{pano}}$ to reinforce spatial geometric
alignment, then projects through two linear layers to the feature dimension consistent with the LRM encoder output tokens, yielding tokens that can be directly fed into the 3DGS Decoder. Compared with
decoding back to RGB images and re-encoding, operating directly in the latent domain preserves the cross-view structural information while reducing unnecessary computational redundancy.

To stabilize this cross-model feature conversion, we adopt a distillation training strategy. The teacher branch takes the actual panoramas and ray geometry as input and produces teacher tokens $T_t$ via the
original LRM encoder; the student branch starts from PanoDiT hidden features and predicts student tokens $T_s$ through the Latent Feature Adapter. The training objective is a bridge distillation loss:
\[
\begin{aligned}
\mathcal{L}_{\mathrm{bridge}}
=
\frac{1}{N}
\sum_{i=1}^{N}
w_i
\Bigl(
&\mathcal{L}_{\mathrm{smoothL1}}(T_s^i, T_t^i) \\
&+
\lambda_{\mathrm{cos}}
\Bigl(
1 -
\frac{
\langle T_s^i, T_t^i\rangle
}{
\|T_s^i\|_2 \|T_t^i\|_2
}
\Bigr)
\Bigr).
\end{aligned}
\]
where $\lambda{\mathrm{cos}} = 0.25$, and target-view tokens are assigned higher weights to focus alignment on the generated content, while source token weights are relatively lower:
\[
w_i =
\begin{cases}
1.0, & i \in \mathcal{T}_{\mathrm{target}},\\
0.25, & i \in \mathcal{T}_{\mathrm{source}}.
\end{cases}
\]
Through distillation training, the output of the Latent Feature Adapter is aligned with the native token distribution of the LRM encoder in terms of statistical distribution, ensuring that the 3DGS Decoder
can stably predict 3DGS from the generated features.

\subsection{Gaussian Decoder}

The Gaussian Decoder is built upon a pretrained panoramic LRM~\cite{jia2026panoworld}. The panoramic LRM is a feed-forward scene reconstruction model whose encoder tokenizes each input panorama into a spatial grid, attaches the
corresponding Plücker ray descriptor to each token, and outputs a sequence of geometry-aware tokens with explicit spatial geometric information. Unlike perspective-image LRMs, the panoramic LRM takes
equirectangular panoramas as input, naturally covering the full visual sphere, and its ray parameterization is well-suited to our panoramic multi-view task setting. The multi-view tokens are aggregated across
views through a transformer with geometric positional encoding to form a unified scene latent, which the Gaussian Decoder then decodes into per-Gaussian parameters including position, opacity, scale, rotation,
and appearance attributes.

Within Pano2World, the panoramic LRM serves two roles: in the preprocessing stage, it takes the source panorama as input to predict the coarse proxy scene $G_0$, providing geometric guidance for subsequent
generation; in the final prediction stage, its Gaussian Decoder receives the tokens output by the Latent Feature Adapter and decodes them into the persistent 3DGS scene $G$. Both stages share the same
pretrained weights, ensuring consistency in system design.

During training, we fine-tune the Gaussian Decoder on top of the LRM's original rendering loss. In addition to the primary rendering supervision, we introduce depth and opacity constraints to suppress floater
artifacts and over-dense Gaussians, and maintain alignment between the Latent Feature Adapter output and the LRM encoder's native token space through a continued bridge distillation loss. We further derive
auxiliary perspective views from the target panoramas to provide additional rendering supervision, encouraging the model to maintain stable local geometry and appearance beyond panoramic rendering. At inference
time, the Latent Feature Adapter fully replaces the original LRM encoder while the rest of the Gaussian Decoder remains unchanged, completing the final 3D scene prediction without rerunning the generation model.

\section{Implementation}

\subsection{Data}

The training data for Pano2World serves two main purposes: learning multi-view panorama completion conditioned on the source panorama and target-pose reference panoramas, and learning the mapping from multi-view panoramic evidence to a 3DGS scene. We therefore use data sources that provide panoramas, camera poses, and geometric supervision as the main training data. We use rendered 3D-FRONT~\cite{fu20213dfront} indoor scenes and RealSee3D~\cite{li2025realsee3d} real-world indoor panorama data as training sources. 

For 3D-FRONT~\cite{fu20213dfront}, we reorganize the original scenes into local source-target training samples to match the local exploration setting of Pano2World. For each indoor scene, we select 1--3 panorama nodes as source panoramas and sample a set of target poses within their nearby spatial neighborhoods. The ground-truth panoramas at the target poses are used as target-view supervision, while the source panoramas and target-view reference panorama serve as input observations.

For RealSee3D~\cite{li2025realsee3d}, the panorama capture locations are fixed, so target views cannot be rendered at arbitrary positions as in synthetic data. We therefore estimate room-level grouping information from the depth and spatial locations of the camera nodes, and construct local training samples within each room. Specifically, we select one panorama node in a room as the source panorama and treat the remaining nodes in the same room as target poses. This allows us to build local camera-translation relationships from real multi-position panorama data, while avoiding layout discontinuities and geometric inconsistency caused by cross-room sampling.

In total, we collect approximately 60K training positions from 3D-FRONT, and further select around 50K positions from RealSee3D for training. Each training sample contains four types of information: source panoramas, target-pose reference panoramas, ground-truth target panoramas, and camera geometry.

\subsection{Training Details}
Training proceeds in three stages. We first fine-tune PanoDiT with LoRA to improve cross-view consistency during multi-view joint denoising. We then freeze PanoDiT and the panoramic LRM and train only
the Latent Feature Adapter, aligning the hidden features of PanoDiT to the native token space of the LRM encoder through bridge distillation. In the third stage, we jointly fine-tune the Latent Feature
Adapter and Gaussian Decoder under end-to-end rendering supervision to further improve final 3D scene prediction quality, while keeping PanoDiT parameters frozen. The three stages are trained on 32, 8, and
8 H200 GPUs respectively, for 10K, 5K, and 5K iterations, with a per-GPU batch size of 1 throughout.

\section{Experiment}

\begin{figure*}[!t]
  \centering
  \includegraphics[width=\textwidth]{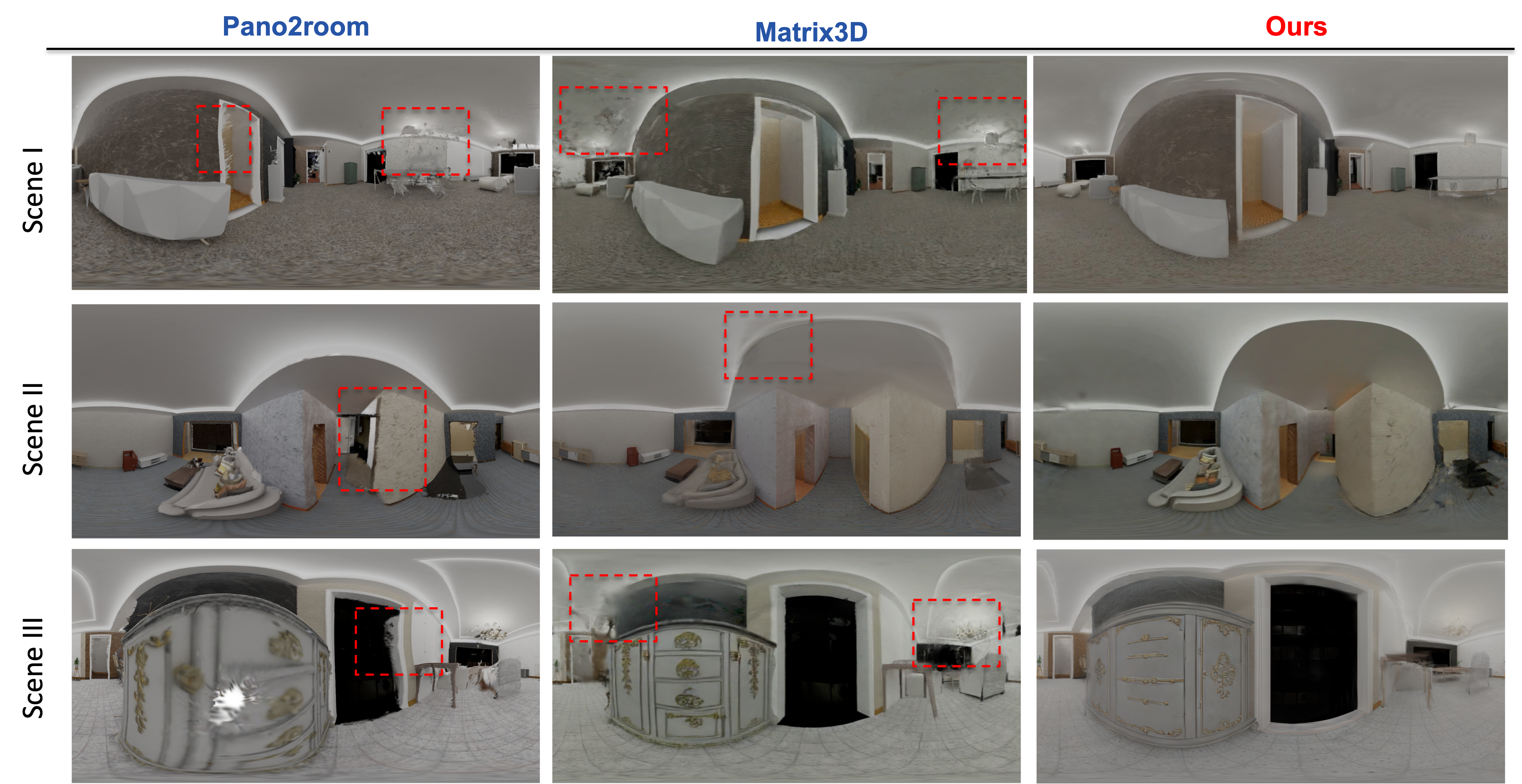}
  \caption{Qualitative comparison on target poses. Additional video comparisons between the input panoramas and the corresponding Pano2World-rendered 3D scenes are provided in the supplementary material.}
  \label{fig:qwen-mvp-case}
\end{figure*}
\subsection{Evaluation Data and Metrics}

\textbf{End-to-end novel-view rendering.}
This protocol directly measures the overall quality of the pipeline from a single source panorama to the final 3DGS scene.
Since this evaluation requires rendering at arbitrary target poses and also needs depth information to align camera scales across different methods, we construct the test set from held-out 3D-FRONT scenes that are not used during training.
Specifically, we select 20 source panorama locations from these unseen scenes and sample target poses within their local neighborhoods for evaluation.
Given a source panorama $I_s$, the model predicts a 3DGS scene $G_{\mathrm{pred}}$, which is rendered at held-out target poses $\{P_k\}_{k=1}^{K}$ to obtain
$\hat{I}_k = \mathrm{Render}(G_{\mathrm{pred}}, P_k)$.
Image reconstruction metrics are then computed against the corresponding ground-truth target panoramas $I_k^{\mathrm{gt}}$.

\textbf{Generated-view 3D consistency.}
End-to-end metrics compare generated results directly against ground-truth target views; however, due to the inherent stochasticity of the generation process, the gap between generated results and ground truth does not solely reflect whether the multiple target views generated by PanoDiT are geometrically and appearance-consistent with one another.
Therefore, we further design a generated-view 3D consistency evaluation protocol and evaluate it on test locations sampled from both 3D-FRONT and RealSee3D.
Specifically, we select 20 source panorama locations from 3D-FRONT and 20 source panorama locations from RealSee3D, generate the corresponding target views, and evaluate whether these generated views can be explained by a single shared 3D representation.
If a set of generated views is mutually consistent in geometry and appearance, a frozen panoramic LRM should be able to reconstruct a unified 3DGS scene from these views and re-render results close to the generated inputs at the same poses.
Conversely, if the generated views exhibit significant geometric drift or texture inconsistency, a single 3D representation will struggle to simultaneously explain all images, and the re-rendering error will increase accordingly.
Specifically, given generated views $\{\tilde{I}_i\}_{i=1}^{N}$, they are fed into a frozen panoramic LRM to reconstruct a scene $G_{\mathrm{LRM}}$, which is then re-rendered at the same poses and compared with the generated inputs.

Both protocols adopt PSNR, SSIM, and LPIPS~\cite{zhang2018lpips} as evaluation metrics. For end-to-end novel-view rendering, the reference is the ground-truth target panorama; for generated-view 3D consistency, the reference is the target panorama generated by PanoDiT.

\begin{table}[t]
\centering
\caption{End-to-end novel-view rendering evaluation. Given a single source panorama, each method renders at held-out target poses and compares against ground-truth target panoramas. Higher PSNR/SSIM and lower LPIPS are better. Runtime is measured on a single A100 GPU at resolution 1024*512 with 8 target poses, including target-view generation and 3DGS decoding}
\label{tab:e2e_nvs}
\begin{tabular}{lcccc}
\toprule
Method &
PSNR$\uparrow$ &
SSIM$\uparrow$ &
LPIPS$\downarrow$ &
Time(s)$\downarrow$ \\
\midrule
Pano2Room~\cite{pu2024pano2room}     & 14.51 &0.587 & 0.566 & 802 \\
HY-World$^\ast$~\cite{hyworld2026hyworld2}   & 15.93 & 0.462 & 0.680 & 1636 \\
PanoLRM~\cite{jia2026panoworld} &  16.05 &  0.728 & 0.325 & \textbf{11} \\
Matrix3D~\cite{yang2025matrix3d}       & \underline{21.54}  & \underline{0.729} & \underline{0.276} & 7487 \\
\midrule
\textbf{Pano2World}           & \textbf{23.56} & \textbf{0.824} & \textbf{0.230} & \underline{197} \\
\bottomrule
\end{tabular}
\end{table}

\subsection{End-to-End Novel-View Rendering}

Table~\ref{tab:e2e_nvs} reports the quantitative results of end-to-end novel-view rendering. To eliminate bias introduced by depth estimation, we uniformly rescale the camera trajectories of all methods to align with the true scene depth. As shown, Pano2World outperforms all baselines by a significant margin across every metric, achieving a PSNR of 23.56. Among the baselines, Matrix3D~\cite{yang2025matrix3d} achieves the closest perceptual quality to ours, benefiting from the strong consistency prior of its large-scale video diffusion model and well-curated training data; however, the enormous inference cost of the 14B video generation model results in a per-scene runtime of 7,487 seconds, $38\times$ slower than Pano2World. Furthermore, the trajectory constraints of video generation models limit their ability to expand in multiple directions simultaneously, leaving Matrix3D 2.02 dB below Pano2World in PSNR. Pano2Room~\cite{pu2024pano2room} is limited to a PSNR of 14.51, primarily because its iterative generation pipeline progressively accumulates errors, which degrades both rendering quality and runtime efficiency. HY-World*~\cite{hyworld2026hyworld2} is a state-of-the-art scene generation method; as its complete implementation for this task is not publicly available, we adapt it by chaining WorldStereo~\cite{zhang2026worldstereo} for video generation and WorldMirror from HY-World 2.0~\cite{hyworld2026hyworld2} for 3DGS reconstruction. Due to the lack of cross-view consistency constraints across independently generated cube-face trajectories, the stitched panoramas exhibit visible color and geometric discontinuities at face boundaries, resulting in the lowest SSIM of 0.46 among all methods. We additionally evaluate the panoramic LRM used in our pipeline in isolation, where it achieves a PSNR of 16.05 using only single-image feed-forward inference, demonstrating the substantial gain brought by our multi-view completion stage.

\begin{table}[t]
\centering
\scriptsize
\setlength{\tabcolsep}{2.5pt}
\caption{Ablation study of PanoDiT major components. ReDir., Perc., Rout., and Cons. denote ReDirector camera conditioning, image-domain perceptual loss, view-aware attention routing, and cross-view consistency constraint loss, respectively. PFLOPs denotes the computation of one PanoDiT forward pass measured in peta floating-point operations.}
\label{tab:ablation}
\resizebox{\columnwidth}{!}{
\begin{tabular}{lcccccccc}
\toprule
Model & ReDir. & Perc. & Rout. & Cons. & PSNR$\uparrow$ & SSIM$\uparrow$ & LPIPS$\downarrow$ & PFLOPs$\downarrow$ \\
\midrule
Base & -- & -- & -- & -- & 19.79 & 0.673 & 0.358 &2.02  \\
A    & \checkmark & -- & -- & -- & 21.73 & 0.790 & 0.294 & 2.03 \\
B    & \checkmark & \checkmark & -- & -- & 27.31 & 0.858 & \textbf{0.197} & 2.03 \\
C    & \checkmark & \checkmark & \checkmark & -- & \underline{27.47} & \underline{0.861} & \underline{0.199} & \underline{0.92} \\
Full & \checkmark & \checkmark & \checkmark & \checkmark & \textbf{29.96} & \textbf{0.891} & 0.201 & \textbf{0.92} \\
\bottomrule
\end{tabular}}
\end{table}

\begin{table}[t]
\centering
\small
\setlength{\tabcolsep}{5pt}
\caption{
Ablation study of the Latent Feature Adapter.
RGB Re-encode reconstructs 3DGS from decoded target panoramas using the original LRM encoder.
Direct Projection directly maps PanoDiT hidden features to the decoder input space, while Full Adapter uses the complete latent feature bridging module.
}
\label{tab:adapter_ablation}
\begin{tabular}{lccccccc}
\toprule
Model & PSNR$\uparrow$ & SSIM$\uparrow$ & LPIPS$\downarrow$ & Time(s)$\downarrow$\\
\midrule
RGB Re-encode  & 22.53   & 0.814   & 0.246 & 231 \\
Direct Projection     &  \underline{22.37} & \underline{0.811} & \underline{0.239}      & \underline{197} \\
Full Adapter     & \textbf{23.56} & \textbf{0.824} & \textbf{0.230}   & \textbf{197} \\
\bottomrule
\end{tabular}
\end{table}

\paragraph{Ablation on PanoDiT.}
We conduct an ablation study to analyze the effects of four key designs in PanoDiT, including ReDirector camera conditioning (R), image-domain perceptual loss (P), View-Aware Attention Routing (V), and cross-view consistency constraint (C), using the generated-view 3D consistency evaluation protocol. As shown in Table~\ref{tab:ablation}, the Base model simply concatenates tokens from all views along the sequence dimension and feeds them into Qwen-Edit for unified denoising. Without explicit 3D geometric constraints, different target views tend to exhibit inconsistent structures and appearances. Introducing ReDirector camera conditioning clearly improves the reconstruction quality, demonstrating the importance of injecting camera poses and panoramic ray geometry into the diffusion model for target-view generation. Building on this, adding the image-domain perceptual loss further improves the metrics, indicating that perceptual supervision in the decoded image space helps preserve local appearance details. The main purpose of View-Aware Attention Routing is to reduce redundant token interactions during multi-view joint denoising; while maintaining comparable generation quality, it reduces the attention computation by about 54.7\% on average, thereby improving the scalability of multi-target-view generation. Finally, the cross-view consistency constraint further improves PSNR and SSIM, suggesting that it promotes geometric coordination among different target views and provides more consistent multi-view observations for subsequent unified 3DGS prediction.

\paragraph{Ablation on Latent Feature Adapter.}
To validate the role of the Latent Feature Adapter, we compare three connection strategies from PanoDiT outputs to the 3DGS decoder.
As shown in Table~\ref{tab:adapter_ablation}, RGB Re-encode first decodes the generated target panoramas into RGB space and then feeds them into the original LRM encoder.
Although this strategy can reconstruct a 3DGS, it introduces additional decoding and re-encoding steps, weakens the cross-view structural information formed during joint denoising, and leads to higher inference time.
Direct Projection directly projects PanoDiT hidden features, avoiding RGB re-encoding; however, without sufficient alignment to the native LRM token space, its performance remains below the full method.
In contrast, Full Adapter achieves the best PSNR, SSIM, and LPIPS without increasing inference time, demonstrating that the Latent Feature Adapter effectively bridges diffusion hidden features and the Gaussian Decoder, thereby improving the reconstruction quality of the final 3DGS.

\subsection{Qualitative Comparison}

Figure~\ref{fig:qwen-mvp-case} shows qualitative comparisons at target poses.
As highlighted by the red boxes, Pano2Room is limited by its iterative per-view completion and geometry update pipeline, which easily leads to local geometric distortion and accumulated errors.
Matrix3D relies on video generation along a continuous trajectory; when target poses expand in multiple directions and deviate from its trajectory distribution, large structural regions such as walls and ceilings tend to exhibit noticeable artifacts.
In contrast, Pano2World benefits from geometrically aligned reference panoramas and joint multi-view denoising, producing more complete and structurally consistent target-view results.

\section{Conclusion}

We present Pano2World, an end-to-end framework that generates a locally explorable 3DGS scene from a single indoor panorama. Pano2World reformulates single-panorama scene expansion as a joint multi-view generation problem over multiple target poses, and improves consistency across target views through geometrically aligned reference panoramas, View-Aware Attention Routing, and ReDirector camera conditioning. Furthermore, we introduce a Latent Feature Adapter that directly transforms the hidden features formed during PanoDiT joint denoising into geometry-aware representations compatible with the Gaussian Decoder, which are then decoded into the final 3DGS. This design avoids the information loss caused by decoding back to RGB space and reconstructing the scene again. Experimental results show that Pano2World achieves better performance in novel-view rendering quality, cross-view consistency, and generation efficiency. Although our experiments mainly use a single source panorama as input, the joint sequence modeling formulation of Pano2World can naturally extend to multiple source panoramas, enabling richer scene coverage and appearance cues to further constrain target-view generation.


\end{document}